\documentclass[review]{elsarticle}

\usepackage{lineno,hyperref}
\usepackage{multirow}
\usepackage{bm}
\usepackage{amsmath,amssymb,amsfonts}
\usepackage{diagbox}
\usepackage[table,xcdraw]{xcolor}
\usepackage{adjustbox}
\usepackage{setspace}
\usepackage{float}

\usepackage{graphicx}
 \usepackage{tabularx}

\usepackage{csvsimple}                       %
\usepackage{float}                           %
\usepackage{longtable}                       %
\usepackage{rotating}                        %
\usepackage{subcaption}                      %
\usepackage{amsmath}

\makeatletter
\def\@author#1{\g@addto@macro\elsauthors{\normalsize%
    \def\baselinestretch{1}%
    \upshape\authorsep#1\unskip\textsuperscript{%
      \ifx\@fnmark\@empty\else\unskip\sep\@fnmark\let\sep=,\fi
      \ifx\@corref\@empty\else\unskip\sep\@corref\let\sep=,\fi
      }%
    \def\authorsep{\unskip,\space}%
    \global\let\@fnmark\@empty
    \global\let\@corref\@empty  
    \global\let\sep\@empty}%
    \@eadauthor={#1}
}
\makeatother

\modulolinenumbers[1]

\journal{Journal of \LaTeX\ Templates}

\begin{document}

\begin{frontmatter}

\title{Segmentation of Weakly Visible Environmental Microorganism Images Using Pair-wise Deep Learning Features \tnoteref{mytitlenote}}
\tnotetext[mytitlenote]{}

\author[myaddress]{Frank Kulwa}
 
\author[myaddress]{Chen Li \corref{cor1Author}}
\ead{lichen201096@hotmail.com}
\cortext[cor1Author]{Corresponding author}

\author[Aaddress2]{Marcin Grzegorzek}

\author[myaddress]{Md Mamunur Rahaman}

\author[Aaddress3]{Kimiaki Shirahama}

\author[Aaddress4]{Sergey Kosov}

\address[myaddress]{Microscopic Image and Medical Image Analysis Group, MBIE College, Northeastern University, Shenyang 110169, PR China}

\address[Aaddress2]{Institute for Medical Informatics, University of Lübeck, Ratzeburger Allee 160, 23538 Lübeck, Germany}

\address[Aaddress3]{Faculty of Science and Engineering, Kindai University, Kindai, Japan}

\address[Aaddress4]{Faculty of Data Engineering, Jacobs University Bremen, Bremen, Germany}

\begin{abstract}
The use of \emph{Environmental Microorganisms} (EMs) offers a highly efficient, low cost and harmless remedy to environmental pollution, by monitoring and decomposing of pollutants. This relies on how the EMs are correctly segmented and identified. With the aim of enhancing the segmentation of weakly visible EM images which are transparent, noisy and have low contrast, a \emph{Pairwise Deep Learning Feature Network} (PDLF-Net) is proposed in this study. The use of PDLFs enables the network to focus more on the foreground (EMs) by concatenating the pairwise deep learning features of each image to different blocks of the base model SegNet. Leveraging  the Shi and Tomas descriptors, we extract each image's deep features on the patches, which are centred at each descriptor using the VGG-16 model. Then, to learn the intermediate characteristics between the descriptors, pairing of the features is performed based on the Delaunay triangulation theorem to form pairwise deep learning features. In this experiment, the PDLF-Net achieves outstanding segmentation results of 89.24\%, 63.20\%, 77.27\%, 35.15\%,  89.72\%, 91.44\% and 89.30\% on the accuracy, IoU, Dice, VOE, sensitivity, precision and specificity, respectively.  
\end{abstract}


\end{frontmatter}

\section{Introduction}
\label{s:int}
The large scale of industrialization and urbanization is providing good living conditions for human beings. However, it has brought serious environmental pollution, including water, air and soil pollution \cite{refer_01}, which raises the risk of diseases such as lung cancer. To eliminate such pollution (pollutants), the use of environmental microbiological method offers higher efficiency, lower cost and harmless compared to the use of chemical methods. It involves the use of  \emph{Environmental Microorganisms} (EMs) for monitoring, controlling and decomposing pollutants. For example, \textit{Epistylis} is employed as a sign of poor quality of water and \textit{Actinophrys} is used for decomposition of organic wastes in sludges \cite{refer_02}. Thus, identification of proper EMs and their corresponding physiological characteristics is necessary. Generally, there are four methods used for identification of EMs. First is the chemical method, which is accurate, but it creates secondary pollution of chemical reagents \cite{refer_03}. Second is the physical method, which requires expensive equipment \cite{refer_03}. The third is the molecular biological method, which distinguishes EMs by sequence analysis of genome \cite{refer_04}. This method needs expensive equipment, is time consuming and requires professional researchers. Fourth is the morphological method, which needs an experienced operator to observe EMs under a microscope and give identification by shape characteristics \cite{refer_05}, \cite{refer_02}. This approach is laborious, time-consuming, inconsistent, and subject to the moods of the operator.

\begin{figure}[H]
\includegraphics[scale=0.557]{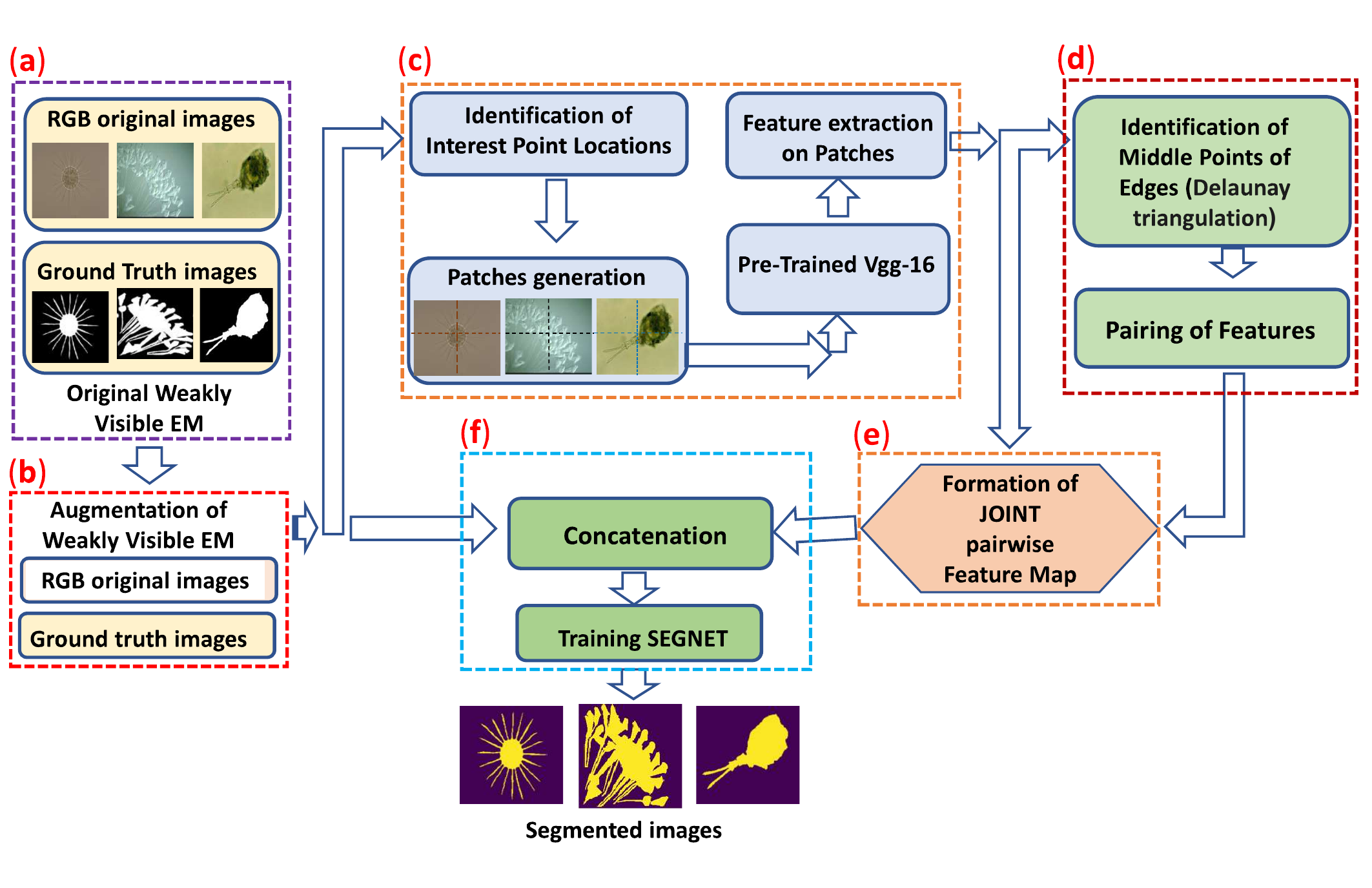}
\caption{The work flow of the proposed PDLF-Net.}
\label{workflow}
\end{figure}

In order to eliminate such drawbacks, automatic image processing techniques are used for the identification of EMs. Image segmentation is a crucial stage in feature extraction \cite{refer_06} and classification \cite{refer_67}, so  we develop a system for segmentation of EM images. The majority of EM samples are obtained from complex environments where large amount of impurities like rubbish is present, which leads to noisy image problems. Moreover, some essential EMs have transparent like body features such as \textit{Ceratium} and \textit{Actinophrys}. This renders less information of the foreground for segmentation tasks, which leads to under-segmentation and poor segmentation results. Furthermore, some EM images suffer from low contrast between the foreground and background, such as \textit{Codosiga} and \textit{Vorticella}, which leads to poor segmentation results. To jointly overcome all segmentation challenges above, we use \emph{Pairwise Deep Learning Features} (PDLFs) concatenated on the convolutional network. The \emph{Pairwise Deep Learning Feature Network} (PDLF-Net) work flow is shown in Figure \ref{workflow}. 

The steps shown in Figure \ref{workflow} from (a) to (f) respectively, are described below. (a) Weakly visible classes: In this study we use an in-house dataset which is is also publicly available in \cite{refer_69} and published in \cite{refer_68}. It contains 21 classes of EMs. Therefore, from 21 classes, the eight most weakly visible classes are selected.    \\(b) Data augmentation: To increase the dataset for training the proposed  CNN, augmentation is performed on both original weakly visible dataset and their corresponding ground truth images.
(c) Feature extraction: Firstly, Shi and Tomas interest points' locations are identified on each image, then all images are meshed into patches (of size $40\times40$ pixels) which are centred at interest points. Then, deep learning features are extracted from each patch using the pre-trained VGG-16 (which is pre-trained on the ImageNeT dataset) and stored. \\(d) Feature pairing: Using the Delaunay triangulation, triangles are identified from interest points, then the middle points of edges of each triangle are identified and used as reference points for pairing the feature maps ( the end of each edge corresponds to extracted features on interest points).
(e) Joint pairwise feature maps formation: The paired features and original features (from interest points) are combined to form a joint pairwise feature for each image. The resultant joint feature map has an average dimension of $46\times1000$ pixels size for each image.
(f) Concatenation and training: At this last stage,  concatenation of the augmented images and their corresponding joint pairwise feature maps are performed at different input stages of the base model (SegNet), to produce the segmented output image.

The contributions of this paper can be folded into three as described below;

1. By extracting deep learning features from small image patches of size $40\times40$ that are centered at the positions of corner interest points, we integrate the  abilities of interest points/ descriptors (hand-crafted features) and deep learning features. The Shi and Tomas theorem is employed to determine the interest points. This allows the network to focus on fine information which is related to edges and corners, thus increasing the segmentation performance and overcome the problem of low contrast and transparency of weakly visible EM.\\
2. Speculating that the middle point between two nearby patches (interest points) have intermediate spatial features, we pair the feature maps of two nearby interest points, to highlight more features around the foreground which could not be learned by base SegNet model. The pairing is achieved using the Delaunay triangular theorem, which concentrates the triangles inside the foreground, thus increasing the focus of the network to learning more foreground which overcome the segmentation challenges in weakly visible EM.\\
3. We concatenate the joint pairwise feature maps to different input scales of the encoder blocks of the base model (SegNet), which generally increase the segmentation results of the network. The joint pairwise feature maps are formed by combining interest point based features and intermediate pairwise features for each image separately.

This paper is organized as follows: Section \ref{s:relatedW} gives a review of related works on microorganisms image segmentation methods (particularly in subsection \ref{s:relatedW1}) while the review on feature extraction and pairwise feature methods are given in subsection \ref{s:relatedW2}. Section \ref{s:method} describes in detail our proposed methods and different key points of our contributions. Experimental results and analysis are discussed in section \ref{s:Exp}. Lastly, conclusion and future works are given in section \ref{s:Conc}.

\section{Related Works}
\label{s:relatedW}
In this section, different works related to our work are reviewed. Section \ref{s:relatedW1} gives a review on segmentation of microorganisms images. Due to the importance of feature extraction in our work, different related works on feature extraction and pairwise features are reviewed in section \ref{s:relatedW2}. Finally, the contributions of our work are given at the end of section \ref{s:relatedW2}.

\subsection{Microorganims Image Segmentation}

\label{s:relatedW1}
Different techniques are implemented to enhance good segmentation performances of microorganisms. These techniques can be categorized into classical and machine learning based techniques \cite{refer_01}. Table \ref{T:segmentationTec}, gives a summary of the categories and subcategories of microorganisms image segmentation methods.


\begin{table}[htbp]
  \centering
  \caption{Categories of microorganisms segmentation methods: (ML means Machine Learning)}
    \begin{tabular}{|p{1.5cm}|p{2.1cm}|p{4.1cm}|p{2.6cm}|}
    \hline
    Categories & Subcategories & Specific methods examples & Related works \\
    \hline
    \multirow{3}[0]{*}{Classical } & Threshold & Otsu, adaptive and global&  \cite{refer_13} \\
          & Edge based & Canny, Sobel& \cite{refer_18} \\
          & Region based & Maker watershed & \cite{refer_20}\\
          \hline
    \multirow{2}[0]{*}{ML} & Unsupervised & k-means, SOM   & \cite{refer_23}, \cite{refer_24}\\
          & Supervised & U-net, SVM, VGG-16 & \cite{refer_27}, \cite{refer_28}, \cite{refer_29}\\
          \hline
    \end{tabular}%
  \label{T:segmentationTec}%
\end{table}%

Classical methods are the traditional techniques which have found broad applications. For instance, in \cite{refer_13} outstanding results are achieved by applying Otsu thresholding in the segmentation of floc and filament. In order to enhance shape feature extraction, an active contour method is used in \cite{refer_18} for segmentation of \textit{Rotavirus-A}. A seed watershed algorithm is applied in \cite{refer_20} for segmentation of \textit{Bacillus subtilis} bacteria in clustered biofilm. Generally, classical methods are associated with challenges such as, they can not work direct on colour images, they need pre-processing like denoising and colour conversion and they cannot perform well on images which have uneven background colours. To overcome above challenges machine learning based methods have been adopted for segmentation.

Machine learning based methods can be categorised into unsupervised and supervised \cite{refer_01}, as shown in table \ref{T:segmentationTec}. Unsupervised machine learning (ML) techniques build their mathematical models from a set of data that contain only input without target output labels (segmentation can be referred to as pixel level classification, in that context the target labels  are the  individual pixel values/ranges in the ground truth mask images. Where, for the case of of unsupervised ML they are not required. An example of unsupervised ML algorithms is the $k$-means clustering). These techniques unsupervisely discover the data pattern and cluster them into segments \cite{refer_22}. For instance, in order to automate the detection of pulmonary tuberculosis (TB) which is caused by  \textit{Mycobacterium tuberculosis}, $k$-means and self organizing map (SOM) clustering were proposed in the segmentation of the basilli from Ziehl-Neelsen sputum smears  \cite{refer_23} and  \cite{refer_24}. While in \cite{refer_26a}, a modified fuzzy divergence clustering method
which is based on Cauchy membership function is leveraged in the segmentation of \textit{Plasmodium vivax} from C channel CMYk color model of images containing the parasites in blood smears. Although unsupervised methods are simple to apply, their ability to learn the pattern of data is  inadequate in transparent images, which is the case for the weakly visible EM.

In recent years the use of supervised methods has shown promising results in segmentation tasks. Supervised machine learning algorithms build mathematical models from a set of labeled data. Example of supervised techniques are convolution neural networks (CNN), support vector machine (SVM) and naive Bayes model. Due to the ability of CNN to capture pattern of data in challenging datasets, they have been used in many works. For instance, \cite{refer_27} increases the receptive field by applying $7\times7$ filter size on fully convolutional network (FCN), this results in an outstanding segmentation performance of 99.7\% accuracy on feline calicivirus images. In \cite{refer_28}, \cite{refer_29}, in order to tackle the challenge of imbalance between the foreground and background, a dice coefficient is applied as a loss function in U-net for segmentation of the rift valley virus and \textit{Leishmania} parasites. To exploit fully the benefits of CNN, a large amount of training dataset is needed. One of the challenges we have in the weakly visible EM is the scarcity of datasets, However  the innovation of strong models such as SegNet \cite{refer_31} and U-net \cite{refer_30} which are capable of working in small number of datasets, gives us a suitable option for our dataset. Moreover, SegNet shows more superiority for having few parameters and hence faster to train, because it passes pooling indeces to the upsampling layers and does not use the heavy deconvolution layers. U-net has been applied in many works for segmentation of EM. Nevertheless, to the best of our knowledge no any work has been done on segmentation of EMs using SegNet, except for one work which uses SegNet directly without any network changes from the original one on sementation of yeast cells \cite{refer_32}. Thus, in this paper we attempt to leverage SegNet for segmentation of weakly visible microorganisms.

\subsection{Feature Extraction and Pairing of Features}
\label{s:relatedW2}
Feature extraction is an important stage in the image processing pipeline. In most cases features are used in image classification and object matching works such as \cite{refer_33}, \cite{refer_34} and \cite{refer_35}. Mainly there are two categories of feature extraction methods, hand crafted and feature learning \cite{refer_36}, as indicated in table \ref{T:FeatureExtraction}. 

\begin{table}[htbp]
  \centering
  \caption{Categories of feature extraction methods}
    \begin{tabular}{|p{2.6cm}|p{5.4cm}|p{2.75cm}|}
    \hline
    Categories & Specific feature (techniques) examples & Related works \\
    \hline
    Hand crafted & Geometric features (Area, perimeter), Local features (SIFT, SURF), Colour, Texture  & \cite{refer_40}, \cite {refer_41},  \cite{refer_42}, \cite{refer_55} \\
    \hline
    Feature Learning & Deep learning (VGG-16, AlexNet, ResNet), BoVW & \cite{refer_46}, \cite{refer_48}, \cite{refer_49}, \cite{refer_53} \\
    \hline
    \end{tabular}%
  \label{T:FeatureExtraction}%
\end{table}%

Hand crafted features are manual features which are extracted based on prior knowledge. For example, color (ie. RGB, HSV, LAB, HUE color modes), texture which is defined by the spatial distribution of pixels in the neighbourhood of an image (ie. energy, entropy, homogeneity, correlation, and contrast \cite{refer_71}) , geometric features (area, perimeter and length), global shape (ie. Krawtchouk moment) and local shape features (ie. SURF and SIFT). Local features are the collection of basic and frequent features that can be used to estimate a class's shape knowledge as they learns from finite samples of training data. Besides, two classes which are fairly similar cannot be distinguished by local features alone. Utilizing  global features convey greater discriminative information of a class domain by making use of more specific and uncommon features \cite{refer_70}. Hand crafted features, particularly local features (SIFT and SURF) are very useful in detection of interest points. Interest points are distinctive spots/regions that help to distinguish between different objects (images). \cite{refer_39}. Corner, blob, and ridge descriptors are examples of interest points. They play an important role in image classification and matching tasks. For example, in \cite{refer_40}, \cite{refer_41} image matching of EMs is achieved used SIFT features, where these features are derived from corner interest points of 10 channels of different color modes.  In \cite{refer_42} edge and Fourier descriptors are applied for classification of EMs using SVM classifier. Interest points (descriptors) are useful in classification and image matching due to the fact that they are invariant to changes of illumination, rotation, and translation. Besides, local discriminant information content is abundant in the local image structure surrounding the interest point \cite{refer_43}. Thus, we leverage the corner descriptors' locations in enhancing the segmentation of weakly visible EM. However, corner descriptors (hand crafted features) are not sufficient to present diverse appearance of weakly visible EM. Therefore, we complement them by using deep learning features (feature learning).

Feature learning (features) are high dimension features generated by the composition of local features such as SIFT. Bag of visual words (BoVW) \cite{refer_37}, sparse coding (which analyse a large number of images to learn set of bases where each expresses a characteristics pattern of a patch \cite{refer_38}), and deep learning features are examples of feature learning \cite{refer_72}. In most cases deep learning features are genereted from training the deep (convolutional) neural networks such as VGG-16, ResNet, and AlexNet.  Deep learning networks represent high level features composed from low level ones. They have superior descriptive power than hand crafted features methods \cite{refer_44}, because they replicate the feature extraction capability of visual cortex in human brain \cite{refer_45}. VGG-16 is among the most superior and used  models in segmentation and classification tasks because of its high  ability in learning features. For example, in \cite{refer_46}, VGG-16 achieves an outstanding performance on classification of viral pneumonia  and bacteria from x-ray images. In \cite{refer_02}, a VGG-16 pre-trained is used as a base model for segmentation in the Deeplab-VGG, this is achieved by replacing the fully connected layers with average pooling, three convolutions and interpolation layer, then use it for initial segmentation of EMs. Leveraging the capability of VGG-16, in this study we employ it in extracting the deep learning features at every location of the detected corner descriptor. 
Because of its robustness and simplicity the Bag of visual words (BoVW) is among the most used feature learning technique. However, because of  the orderless representation of local features in it, it does not achieve maximum performance. To remedy that and improve the performance of BoVW, some studies have considered spatial arrangement of features to discover higher order in BoVW for object matching and classification \cite{refer_50}, \cite{refer_51}. Among the methods of arranging spatial features is by pairing of close visual words \cite{refer_52}. For instance, in \cite{refer_53} and \cite{refer_54} pairing is done on visual words (where Prior to pairing, feature descriptions are mapped to the visual words, and then pairing is carried out on the visual words). Yet, the underlying distribution of pairs of neighboring local feature descriptors appears to be ignored by the pairing of visual words. To address that, \cite{refer_48} and \cite{ refer_49} suggested that the pairing  of spatial close local descriptors (such as SIFT) can be done before the building of BoVW. This seem to achieve maximum improvement on classification of challenging dataset. Motivated by the concept of pairing features and to the best of the authors' knowledge, there is no any work which has been done on pairing of deep learning features for segmentation task, thus in this study we pair deep learning features generated from corner interest points' locations and concatenate them to the base model for segmentation of weakly visible EMs.

\section{Methods}
\label{s:method}
\begin{figure}[H]
 \centering
\begin{subfigure}[t]{0.265\textwidth}
  \includegraphics[width=1\textwidth]{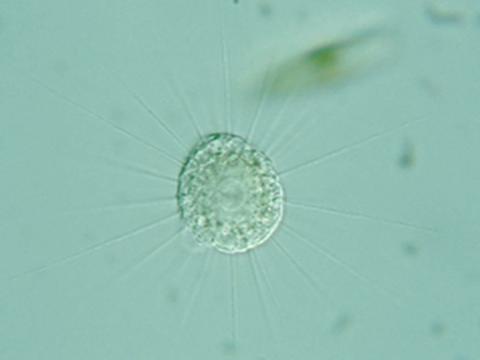}
 \caption{\textit{Actinophrys}}
 \label{fig:results55a}
 \end{subfigure}%
 \hspace{0.032cm}
 \begin{subfigure}[t]{0.265\textwidth}
  \includegraphics[width=1\textwidth]{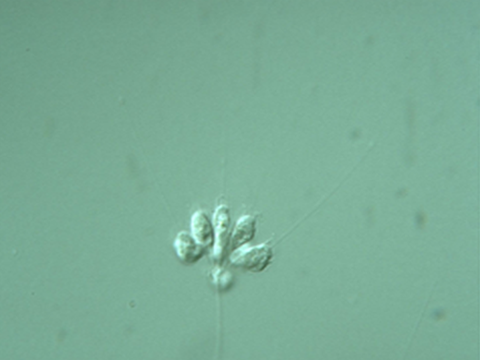}
\caption{\textit{Codosiga}}
 \end{subfigure}
 \hspace{0.00cm}
 \begin{subfigure}[t]{0.265\textwidth}
  \includegraphics[width=1\textwidth]{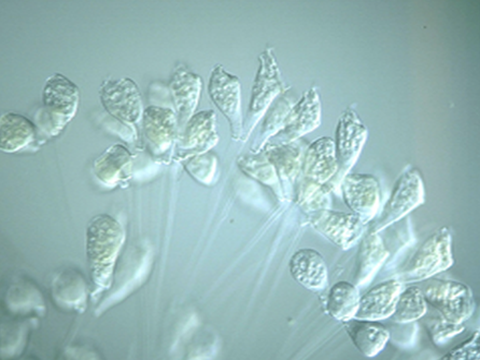}
 \caption{\textit{Epistylis}}
 \end{subfigure}
 
 \begin{subfigure}[t]{0.265\textwidth}
  \includegraphics[width=1\textwidth]{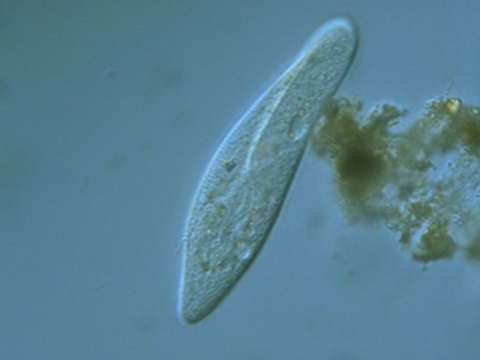}
 \caption{\textit{Paramecium}}
 \label{fig:results55a}
 \end{subfigure}%
 \hspace{0.032cm}
 \begin{subfigure}[t]{0.265\textwidth}
  \includegraphics[width=1\textwidth]{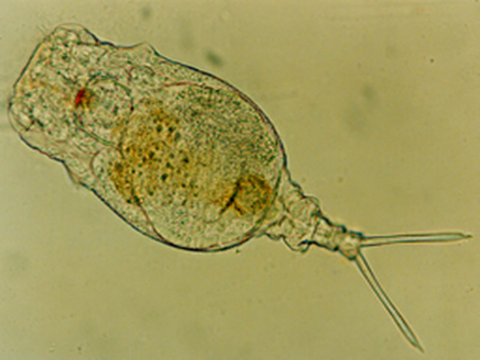}
\caption{\textit{Rotifera}}
 \end{subfigure}
 \hspace{0.00cm}
 \begin{subfigure}[t]{0.265\textwidth}
  \includegraphics[width=1\textwidth]{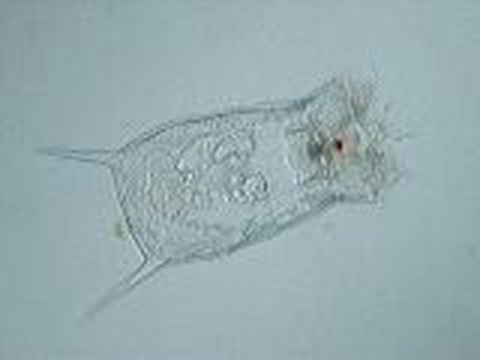}
 \caption{\textit{Keratella Quadrata}}
 \end{subfigure}
\caption{Weakly visible EMs} 
\label{WeaklyEM}
\end{figure}

This section desctribes in details the novel techniques used in this paper. The main focus being on tackling the segmentation challenges on weakly visible EMs. These are EMs which show poor segmentation results in our initial tests using the original base model SegNet. Example of weakly visible EMs are shown on figure \ref{WeaklyEM}.

As observed in figure \ref{WeaklyEM}, weakly visible EMs suffer from low contrast, transparency and  indistinct boundary between background and foreground. To be able to achieve better segmentation results, the following techniques are applied.

\subsection{SegNet}
\label{SegNet}
\begin{figure}[h]
\centering
\includegraphics[scale=0.8]{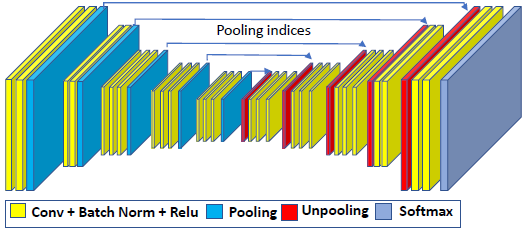}
\caption{SegNet, a base model for the proposed network.}
\label{figureSegNet}
\end{figure}

SegNet is one of the powerful models in computer vision for semantic segmentation \cite{refer_31}. It consists of the encoder and decoder, as shown in figure \ref{figureSegNet}.

The encoder of the SegNet consists of 13 convolutional layers similar to VGG-16, without the last fully connected layers. Thus, the encoder network is largely reduced in parameters compared to VGG-16 and can easily be trained. Each of the 13 encoder layers constitutes of a series of convolutional layer with 64 filter banks (contrary to the original SegNet which use $7\times7$ filter size, we apply $5\times5$  to reduce the receptive field suitable for small size of EMs in images). Followed by batch normalization. Then a ReLu activation function $f(x) = max(0,x)$ which eliminates negative values follows. To achieve translation invariance over small spatial shift of input images, max-pooling with window size of $2\times2$ and stride 2 (non overlapping window) follows, which results into output being sub-sampled by the factor of 2 after each step. The application of 13 max-pooling down sampling layers in the encoder achieves more robust pixel level classification but there is a loss in spatial resolution of feature maps (boundary details). To overcome this, the boundary information in the encoder feature maps are captured and stored before next sub-sampling in each stage by storing the max-pooling indices which are more efficient for restoring boundary information and require less memory. The decoder network (which has similar convolution layers in up sampling manner) upsamples the input feature maps using the memorized max-pooling indices from corresponding encoder feature maps. Each upsampling is followed by convolution and batch normalization layer to produce dense features that are similar in size to the corresponding inputs at the encoder. Finally, the softmax is used as the classification layer. We utilize SegNet as the base model for binary segmentation of weakly visible EMs. For all experiments we use binary cross entropy as a loss function and SGD optimizer with learning rate of 0.01 and momentum of 0.9. Although ReLu has shown some drawbacks such as decreasing the performance in the gradient descent operations because all gradient values would be zero when the activation values are zero \cite{refer_66}, we still opt to use it instead of LeakyReLu which provides effective learning even when the values of activation are zero. This is because during our preliminary experiments on activation functions, the average results for ReLu were slightly highter than LeakyReLu by the margin of 0.19\% accuracy.

\subsection{Feature Extraction}
\label{featureExt}
Due to the challenges on weakly visible EMs dataset, the base model misses fine information from images during training, which gives poor segmentation results when using SegNet alone. Therefore, we use external pairwise features to enhance the performance of the base model by combining the advantage of interest points' locations (hand crafted features) and deep learning features. Specific techniques are describes below;

\subsubsection{Shi and Tomas Intest Points' Location}
In order to enhance the segmentation results, we choose to use corner interest points, because from test/initial experiments the base model misses tiny outer corners and boundaries on the weakly visible EMs due to low contrast and transparency on images. A corner is as a place or point in the image where a small change in location causes a significant change in intensity in both the horizontal (X) and vertical (Y) axes. It can also be described as the intersection of points on an object's contour edges that preserve significant object's features \cite{refer_58}. Shi and Tomas corners theorem is one the most superior corner theorems \cite{refer_60}. Simply the Shi and Tomas theorem operates on three steps; 

Firstly, it is to find the window which produces high variation in intensity with a small change in the $X$ and $Y$-axis. Numerically, to find a window that can produce large variation, let the window be centred at $(x,y)$ and an intensity at this point be $I(x,y)$. $I(x,y)$ is an individual intensity at a position which varies from 0 to 255 for gray level image. When the window is shifted by $(u,v)$, the intensity at the new location will be $I(x+u,y+v)$ and $[I(x+u,y+v)-I(x,y)]$ is the difference in intensity due to shift. For a corner, this difference must be high. Therefore we maximize this term  by differentiating it with respect to $x$ and $y$.  Letting $w(x,y)$ be the weights of pixels over the rectangular or a Gaussian window, Then, $E(u,v)$ which is the difference between the original and the shifted window, is defined as : 

\begin{scriptsize}
\begin{equation}
\begin{aligned}
E(u,v)= \displaystyle\sum_{x,y}^{} w(x,y)[I(x+u,y+v)-I(x,y)]^2
\end{aligned}
\end{equation}
\end{scriptsize}

Applying the Taylor series with only the first order, which is
\begin{scriptsize}
\begin{equation}
\begin{aligned}
T(x,y)=f(u,v)+(x-u)f_x(u,v) + (y-v)f_y(u,v)..
\end{aligned}
\end{equation}
\end{scriptsize}
Rewritting the shifted intensity using the above formula:
\begin{scriptsize}
\begin{equation}
\begin{aligned}
I(x+u, y+v)=I(x,y)+ \frac{\mathrm d (x,y)}{\mathrm d x} \left (u \right) + \frac{\mathrm d (x,y)}{\mathrm d y} \left  (v\right) \\\
\end{aligned}
\end{equation}
\end{scriptsize}
\begin{center}
\begin{scriptsize}
Let:$\frac{\mathrm d (x,y)}{\mathrm d x} =I_x$ , and, $\frac{\mathrm d (x,y)}{\mathrm d y}=I_y$
\end{scriptsize}
\end{center}
$I_x$ and $I_y$ are image derivatives in X and Y directions respectively.
Then,
\begin{scriptsize}
\begin{center}
$E(u,v)= \displaystyle\sum_{x,y}^{} w(x,y)[I(x,y)+I_xu+I_yv-I(x,y)]^2$\\
\end{center}
\end{scriptsize}

\begin{scriptsize}
\begin{equation}
\begin{aligned}
E(u,v)= \displaystyle\sum_{x,y}^{} w(x,y)[I_xu+I_yv]^2
\end{aligned}
\end{equation}
\end{scriptsize}
Expanding the above equation,\\
\begin{scriptsize}
\begin{equation}
\begin{aligned}
E(u,v)= \displaystyle\sum_{x,y}^{} w(x,y)[I^2_xu^2+I^2_yv^2+2I_xI_yuv]
\end{aligned}
\end{equation}
\end{scriptsize}
Taking u,v out and rewritting in matrix notation, the equation becomes;

\begin{scriptsize}
\begin{equation}
\begin{aligned}
E(u,v)=(u,v)M\begin{pmatrix} u\\v  \end{pmatrix}\\
\end{aligned} 
\end{equation}
\end{scriptsize}
where,
\begin{scriptsize}
\begin{center}
$M=w(x,y)\begin{pmatrix} \displaystyle\sum_{x,y}^{} I^2_x & \displaystyle\sum_{x,y}^{} I_xI_y\\\displaystyle\sum_{x,y}^{} I_xI_y & \displaystyle\sum_{x,y}^{} I^2_y \end{pmatrix}$
 \end{center} 
\end{scriptsize}
Where, $M$ is a symmetric $2\times2$ matrix whose eigenvalues are used to determine whether the scanned window contains a corner.

Secondly, Calculating the score value $S$ associated with  scanned window \cite{refer_60}. It is given by;
\begin{scriptsize}
\begin{equation}
S=min(\lambda_1,\lambda_2)
\end{equation}
\end{scriptsize}
where, $\lambda_1$ and $\lambda_2$ are eigenvalues of the matrix $M$.

Thirdly, is to determine points along the shift of the window that can be considered as corners. For the point to be considered as corner, the score value $S$ must be greater than the specified value (if both the $\lambda_1$ and $\lambda_2$ are greater than the minimum threshold values respectively).

\begin{figure}[H]
 \centering
\begin{subfigure}[t]{0.265\textwidth}
  \includegraphics[width=1\textwidth]{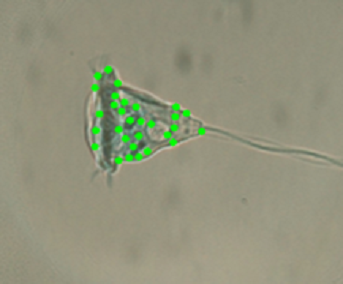}
 \caption{\textit{Vorticella}}
 \label{fig:results55a}
 \end{subfigure}%
 \hspace{0.032cm}
 \begin{subfigure}[t]{0.265\textwidth}
  \includegraphics[width=1\textwidth]{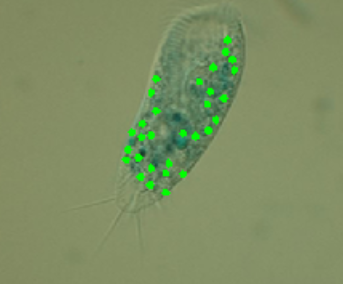}
\caption{\textit{Stylongchia}}
 \end{subfigure}
 \hspace{0.00cm}
 \begin{subfigure}[t]{0.265\textwidth}
  \includegraphics[width=1\textwidth]{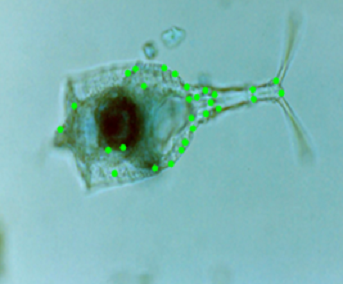}
 \caption{\textit{Rotifera}}
 \end{subfigure}
\caption{Shi and Tomas corner points (in green) detected on weakly visible EMs} 
\label{InterestPoints}
\end{figure}

Shi and Tomas theorem show superiority by having stability, invariant to scale changes, invariant to  translation and invariant to rotation \cite{refer_60}, moreover, comparing with Harris corner points which we applied in our previous work \cite{refer_41}, Shi and Tomas gives better results and more useful interest points than  Harris'. Thus, we use it determine corner points on every image.
Example of images with corner points indicated on them are shown in figure \ref{InterestPoints}.
  
As can be seen from the figure \ref{InterestPoints}, the interest points are capable of identifying corner points that contain unique information about the EMs, which were ignored by the base model (SegNet) during our initial tests for the base model. It should be noted that, in this study  we limit the number of corner points between 10 to 15 (due to computational complexity of the feature extraction model). Then the coordinates of each corner point are identified and stored. We take advantage of the corner points by meshing each image into patches of size $40\times40$ which are centred at each corner points as shown in figure \ref{featureExt2} part (a) and (b).
Then from each patch, we extract deep learning features using convolution neural network VGG-16.

\subsubsection{VGG-16}
\begin{figure}[H]
\centering
\includegraphics[scale=0.8]{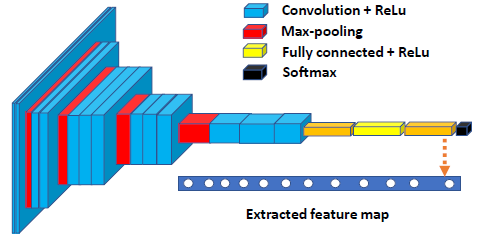}
\caption{VGG-16 network showing feature extraction layer (fully connected layer 3)}
\label{Vgg16_1}
\end{figure}

VGG-16 is a very deep convolution neural network for image recognition, proposed by Simonyan et al in \cite{refer_61}. It is upgraded from AlexNet by replacing large sized kernel filters (11 and 5) with $3\times3$. It has achieved high accuracy in many image classification tasks. It contains 21 layers with only 16 weight layers, which include 13 convolution layers with very small receptive fields of $3\times3$ (which gives its capability to capture the pattern of tiny information fields), followed by max-pooling layers of size $2\times2$ and stride 2 which decreases the spatial resolution of the feature maps. In the end there are three fully connected layers, which combines all learned features from previous layers and generalize them for classification. ReLu activation function is applied to all hidden layers. Lastly is the classifier layer. In order to leverage the fully connected (FC) layers, we extract deep learning features on the last FC layer. The dimension of each extracted feature is about $1\times1000$ pixels size. The figure \ref{Vgg16_1} shows the VGG-16 network layers and the point form which deep learning features are extracted.

\begin{figure}[htbp!]
\centering
\includegraphics[scale=0.7]{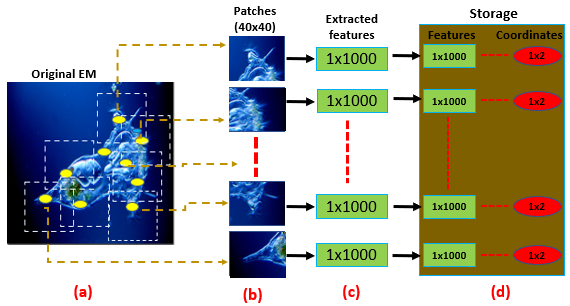}
\caption{The work flow for extraction of deep learning features from EM images (a) detection of interest points' coordinates indicated by yellow colour on the original image, (b) meshing of patches which are centred at interest points (c) deep learning feature extraction (d) storage of features and interest points' coordinates}
\label{featureExt2}
\end{figure}

Due to the small number of weakly visible EM which can not train the VGG-16  from scratch for better results, we use the transfer learning concept to optimize the VGG-16 extracted features.
VGG-16 network, pre-trained on the ImageNet dataset has proven success in many works when fine turned on other datasets for classification \cite{refer_62}. Therefore, we fine tune the pre-trained VGG-16 using weakly EMs and extract deep learning features. For each image, 10 patches of size $40\times40$ are meshed out and from each patch deep learning features are extracted (each patch is centred at interest points' coordinate). Then 10 features for each image are stored parallel to their corresponding interest points' coordinates. Figure \ref{featureExt2} summarizes the process of deep learning features extraction. 
 
\subsection{Feature Pairing}
To pair feature maps which have been extracted from the interest points' coordinates, we use the Delaunay triangulation theorem.

\subsubsection{Delaunay Triangulation (DT) Theorem}
\begin{figure}[h]
 \centering
\begin{subfigure}[t]{0.265\textwidth}
  \includegraphics[width=1\textwidth]{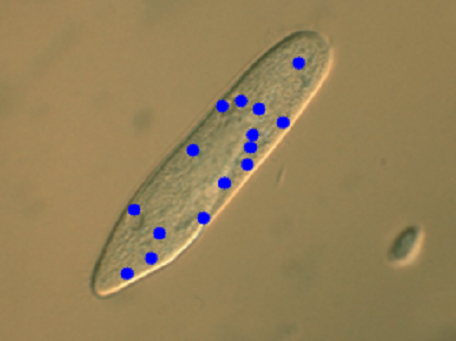}
 \label{fig:results55a}
 \end{subfigure}%
 \hspace{0.032cm}
 \begin{subfigure}[t]{0.265\textwidth}
  \includegraphics[width=1\textwidth]{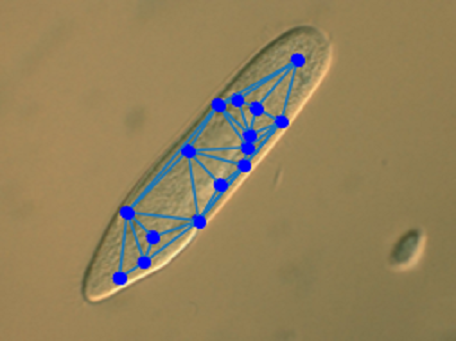}
 \end{subfigure}
 \hspace{0.00cm}
 \begin{subfigure}[t]{0.265\textwidth}
  \includegraphics[width=1\textwidth]{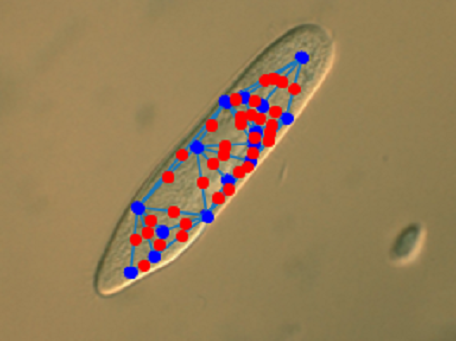}
 \end{subfigure}
 
 \begin{subfigure}[t]{0.265\textwidth}
  \includegraphics[width=1\textwidth]{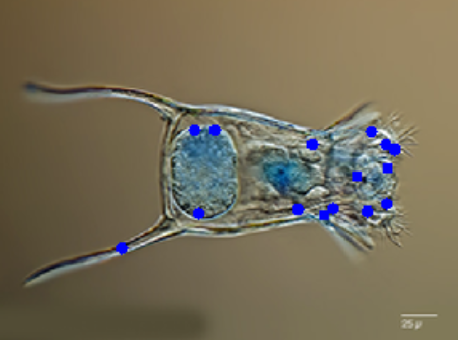}
 \label{fig:results55a}
 \end{subfigure}%
 \hspace{0.032cm}
 \begin{subfigure}[t]{0.265\textwidth}
  \includegraphics[width=1\textwidth]{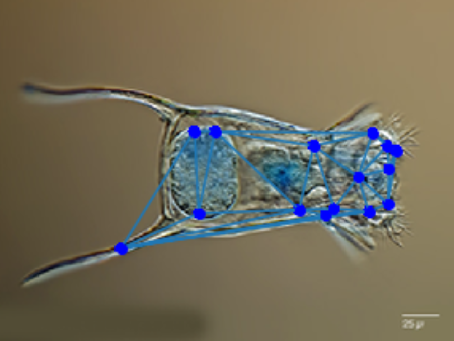}
 \end{subfigure}
 \hspace{0.00cm}
 \begin{subfigure}[t]{0.265\textwidth}
  \includegraphics[width=1\textwidth]{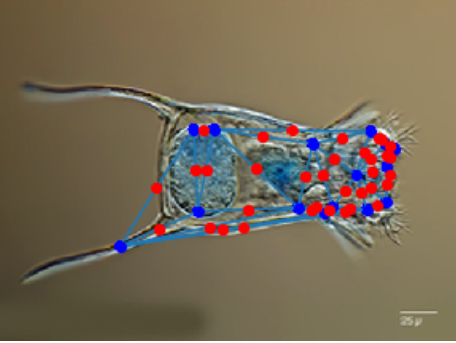}
 \end{subfigure}
 
 \begin{subfigure}[t]{0.265\textwidth}
  \includegraphics[width=1\textwidth]{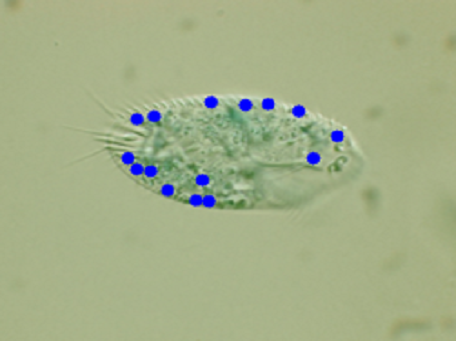}
 \caption{}
 \label{fig:results55a}
 \end{subfigure}%
 \hspace{0.032cm}
 \begin{subfigure}[t]{0.265\textwidth}
  \includegraphics[width=1\textwidth]{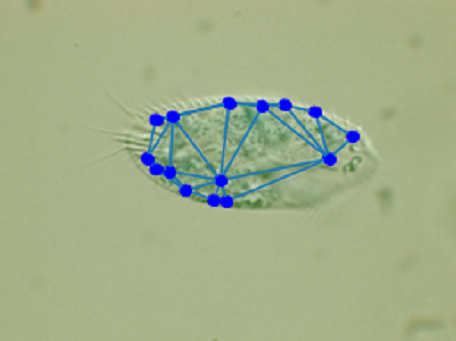}
\caption{}
 \end{subfigure}
 \hspace{0.00cm}
 \begin{subfigure}[t]{0.265\textwidth}
  \includegraphics[width=1\textwidth]{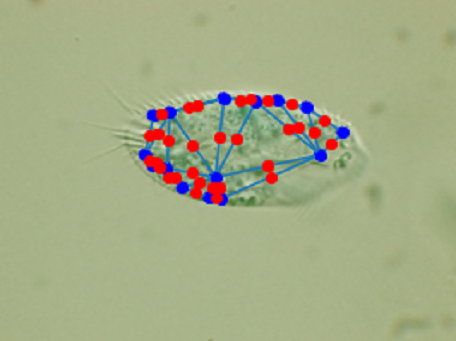}
 \caption{}
 \end{subfigure}
\caption{Pairing of features (a) Detection of interest point coordinates which are indicated in blue colour, (b) application of Delaunay triangulation, (c) pairing of features at middle points of edges which are indicated in red. Blue coloured points are interest points} 
\label{Delaunay}
\end{figure}

DT theorem is one of the most robust graphical theorems for the representation of data. It is the triangulation theorem which forms triangles (Delaunay triangles) by connecting each data (coordinates) to its nearest neighbour, such that the circumcircle associated with each triangle does not contain a point in its interior \cite{refer_63}. Geometrically, Delaunay triangulation for a given set \textbf{A} of discrete data in a plane is a triangulation (DT), such that no data in \textbf{A} is inside the circumcircle of any triangle in DT(\textbf{A}). Delaunay triangulation maximizes the minimum angle of all the angles of the triangles in the triangulation \cite{refer_64}. It is very effective for presentation of scattered data as it concentrates all data inside the major circumcircle formed by the most outer triangle as shown in figure \ref{Delaunay} (b). Due to strong presentation power, it is used in many image matching works \cite{refer_63b}, \cite{refer_65}. Moreover, it is tolerable to spatial displacement of data (image objects)  because it keeps the same association of the nearest objects within the image, regarded that the distortion is uniform all over the image.

The Delaunay triangle edges are formed by connecting nearest neighbour data points. This means two points (vertices) which share the same edge (line) have close related characteristics (features). Thus, the middle point of the edge contains features which are an average of the edge end point features. Although (from our experiments) few middle points might be out of the EM's body which will have non similar characteristics between the edge end points; these points are very few (less than 5\% of all the middle points). More than 95\% of the middle points are within the main body of the EM (foreground) and have intermediate characteristics between the corresponding edge end points as it can be observed in figure \ref{Delaunay} (c). Owing to this, we pair the features which correlates to the vertices sharing same edge, so as to get the features of the middle point of edges. By so doing, we increase  the foreground's influence during segmentation as shown in figure \ref{Delaunay} (c). The pairing of features is done by using the geometric principle of the the middle point of straight line, because the edges of the triangles are straight lines. This is done by averaging the two feature vectors (maps) corresponding to each edge end coordinates as described in the equation \ref{Eq:middlepoint1} and \ref{Eq:PairwiseFeature1}. The edge coordinates are the interest points' coordinates with their corresponding features ($1\times1000$ dimension) extracted from patches.

Let the coordinate of the two end points of an edge be represented by $(X_1 ,Y_1)$ and $(X_2 ,Y_2)$. The corresponding feature maps of the patches centred at these two points be $F_1$ and $F_2$.\\
The middle point coordinate $X_m, Y_m$ is given by;

\begin{scriptsize}
\begin{equation} 
 X_m ,Y_m=\frac{(X_1+X_2)}{2},\frac{(Y_1+Y_2)}{2}
 \label{Eq:middlepoint1}
\end{equation}
\end{scriptsize}

The pairwise feature map $F_m$ which corresponds to middle point $X_m$, $Y_m$ is given by;
\begin{scriptsize}
\begin{equation} 
 F_m=\frac{(F_1+F_2)}{2}
 \label{Eq:PairwiseFeature1}
\end{equation}
\end{scriptsize}

In average 36 to 43 pairwise features ($Fm$) are formed from 10 original features for each image.

\subsection{Joint Pairwise Feature Formation}
At this stage, we join the features formed on the interest points' coordinates ($F1, F2$...) and pairwise features ($Fm$...). The average amount of pairwise features for each image is between 36 and 43. 10 features originate from interest points. Thus, we form the joint feature maps by appending these features vertically. This joining style has shown best results from the tests done during experiments. The average joint feature maps sizes range from $46\times1000$ to $53\times1000$ for different images. Therefore, each joint feature map corresponds to one original image. Because the dominant features are pairwise features, we name the features as joint pairwise features (Pairwise features). After formation of joint features, they are stored parallel to their original images and ground truth images.

\subsection{Concatenation and Training}
\begin{figure}[H]
\centering
\includegraphics[scale=0.55]{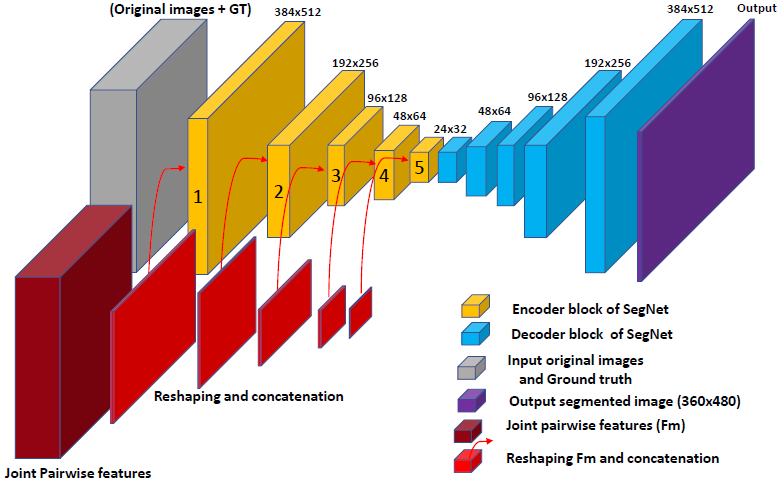}
\caption{The proposed network on concatenation of joint pairwise features (in red) to the base SegNet encoder blocks (in yellow)}
\label{GeneralFigure}
\end{figure}

Both the original images and their corresponding joint pairwise features point to similar ground truth (GT) images. During training, the original images and corresponding ground truth images are fed to the input (first block) of the base model (SegNet). The joint pairwise features are resized at different sizes to fit the spatial dimensions of the encoder blocks of the the SegNet. These dimensions are $384\times512$, $192\times256$, $96\times128$, $48\times64$ and $24\times32$ for first, second, third, fourth and fifth blocks respectively. Then we concatenate the joint pairwise features  at different blocks of the encoder in the SegNet, as shown in figure \ref{GeneralFigure} of the general proposed network.  

We apply different options for concatenating the joint feature maps to the SegNet model. Example of the options are, concatenation at block 1 only, block 2 only, block 3 only, block 1 and 2 only, block 3 and 5 only, block 1, 2 and 5 only.
\section{Experiment}
\label{s:Exp}
\subsection{Experimental Settings}
\subsubsection{Dataset}
During experiments, we use Environmental Microorganism Dataset 5th Version (EMDS-5), which is a newly released version of EMDS. The dataset contains 21 classes of EMs.  However, in this research, we select only 8 classes which show poor performance on the base model SegNet during our initial experiments. We name these images as weakly visible EMs. Particularly, these classes are \textit{Actinophrys} which is denoted as weak data class 1 (DC1), \textit{Codosiga} denoted as DC2, \textit{Epistylis} denoted as DC3, \textit{Paramecium} denoted as DC4 and \textit{Rotifera}, \textit{Vorticella}, \textit{Keratella Quadrala},  \textit{Stylongchia} denoted as DC5, DC6, DC7, DC8 respectively. Each class contains 20 original microscopic images and their corresponding ground truth (GT) images. Therefore, in total there are 160 EMs. It should be noted that every image contains one microorganism in it (not in colonies) except for class DC3 and DC4 where some images contain two microorganisms of the same species. An example of such EMs can be seen in Fig. \ref{segmented images}. True corners of the foreground are most important in this research. Thus, in order to reduce the possibility of false corners we crop all images which have outer highlighted square frames at the edges of the images and remain with only the true background and foreground. Then all images are resized to $360\times480$ pixel sizes so as to fit in the SegNet input layer size. 

\subsubsection{Training, Validation and Testing Dataset}
The dataset is divided into training, validation and testing in ratio 1:1:2 respectively. However, in order to overcome the overfitting due to small dataset and improve the performance of our segmentation models, we applying augmentation on all original weakly visible EM and their corresponding GT images. We augment by rotating them by 90, 180 and 270 degrees, and flipping them vertically and horizontally. This result into 960 images in total while having 30:30:60 images for each class for training, validation and testing respectively. Then from each RGB image joint pairwise feature maps are extracted and distributed into same ratio 30:30:60 corresponding to each class.

\subsubsection{Experimental Environments}
To conduct the experiments, we use a work station with Intel (R) Core(TM) i7-7700 CPU with speed of 3.60Hz. RAM of 32GB and NVIDIA GeForceGTX 1080 8GB. For implementation of the networks, we use python 3 and Keras framework with Tensorfow as backend.

\subsection{Evaluation Metrics}

In order to evaluate quantitatively and compare the segmentation results of  different approaches, we use accuracy (Acc), Dice, intersection of union (IoU, volumetric overlap error (VOE), Sensitivity (Sens), Precision (Prec) and Specificity (Spec). \textbf{Accuracy:} measures the percentage of pixels in an image which are correctly classified. Accuracy and specificity sometimes mislead results on segmentation when the object of interest is small compared to background (which is the case for our dataset). Because these measures are biased mainly on how well negative pixels (background) is predicted. Thus, we use more than one metric for correct analysis of the results. \textbf{Dice coefficient}: also known as  F1 score, is widely used for evaluation of segmentation performance. The definition of Dice is given in table \ref{metrics}. \textbf{Intersection over union}: Also known as Jaccard coefficient, measures the percentage overlap between the target mask and the prediction output. \textbf{Volumetric overlap error}: Is the complement of Jaccard coefficient. Table \ref{metrics}, summaries definition of these metrics.

\begin{table}[htbp]
  \centering
  \caption{Definitions of used metrics}
    \begin{tabular}{|p{0.88cm}|p{4cm}|p{0.88cm}|p{4.7cm}|}
    \hline
    \vspace{0.05cm}
    Metric  & Definition & Metric & Definition \\    
    \hline
    \vspace{0.05cm}
Acc, Spec&$\frac{TP+TN}{TP + FP + TN+ FN}$,$\frac{TN}{TN+FP}$ & Dice&$\frac{2\times| W_{prd} \bigcap W_{gt}|}{|W_{prd}| + |W_{gt}|}$=$\frac{2\times TP}{2\times TP + FP + FN}$ \\
\hline
 \vspace{0.05cm}
IoU, Sens& $\frac{TP}{TP + FP + FN}$, $\frac{TP}{TP + FN}$& VOE, Prec& $\frac{FP + FN}{TP + FP + FN}$, $\frac{TP}{TP + FP}$ \\

\hline
    \end{tabular}%
  \label{metrics}%
\end{table}%

From table \ref{metrics}, $W_{prd}$ represents the predicted foreground by the model. $W_{gt}$ represents the foreground in the ground truth image. During segmentation of  the EM, images are partitioned into two class pixels representing the foreground (the EMs) and the background. True positive (TP): is an outcome when the model correctly predicts the positive class. True negative (TN): is when the model predicts the negative class correctly. False negative (FN): is the outcome when the model predicts negative while it is actually positive. True negative (TN): is when the model predicts negative and it is actually negative. All the evaluation metrics are defined based on these terms TN, TP, FN, FP as shown in table \ref{metrics}. For analysis purposes, the greater the values of accuracy, Dice, IoU, sensitivity, precision and specificity indicate better segmentation results and the smaller the value of VOE indicates better results and vice versa. 

\subsection{Evaluation of the Pairwise Deep Learning Features Network (PDLF-Net) on small Dataset Without Augmentation}
Because the PDLF-Net originates from SegNet, therefore in this section we compare the segmentation performance of the PDLF-Net and SegNet on a small dataset (Each class having 5:5:10 dataset for training, validation and testing respectively). In our initial experiments we examined the performance of the PDLF-Net on different options of concatenating the joint pairwise features to different blocks of the encoder, such as concatenation at one block only of the PDLF-Net encoder as shown in figure \ref{GeneralFigure}, two blocks simultaneously, three blocks simultaneously, four blocks simultaneously and five block simultaneously. Referring to figure \ref{GeneralFigure}, these concatenation options can be described as concatenation at block 1 only, block 2 only, block 3 only, block 4 only, block 5, block 1 and 2 only, block 1 and 3 only following this order up to block 1, 2, 3, 4, and 5 only. We found that the performances are better when the concatenation is only at one block either block 1, block 2, block 3, block 4 and block 5 only. The increase in the number of concatenation blocks simultaneously leads to over-segmentation. Thus, we focus our research on concatenation at one block for all other experiments which we present in this paper.
We compare and examine the performance of the PDLF-Net on small dataset of weakly visible classes by treating each class alone. Table \ref{T1:withoutAug} shows the performance of the PDLF-Net with concatenation at different blocks and the original SegNet.


\begin{table}[htbp]
\begin{scriptsize}

  \centering
  \caption{Segmentation results of the original SegNet and PDLF-Net while concatenation is at block 2 and 5, on evaluation small dataset. The results for each weakly visible class 1 to 8 (DC1 - DC8) are given separately in each row. The evaluation metrics are accuracy (Acc), intersection of union (IoU), Dice and volumetric overlap error (VOE), sensitivity (Sens), precision (Prec) and  specificity (Spec). Red coloured values are maximum/best achieved values in each metric}
        \begin{tabular}{|p{0.38cm}|p{0.35cm}p{0.35cm}p{0.35cm}p{0.35cm}p{0.35cm}p{0.35cm}p{0.46cm} |p{0.35cm}p{0.35cm}p{0.35cm}p{0.35cm}p{0.35cm}p{0.35cm}p{0.5cm}|}
        \hline
      & \multicolumn{7}{c}{SegNet  [\%]} \vline& \multicolumn{7}{c}{Block 2 [\%]}\vline \\
      \hline
    Data & IoU & Dice & VOE & Sens & Prec & Spec & Acc & IoU & Dice & VOE & Sens & Prec & Spec & Acc \\
    DC1 & 42.10 & 59.25 & 57.91 & 50.17 & 90.80 & 50.00 & 72.55 & \textcolor[rgb]{ 1,  0,  0}{45.00} & 62.05 & \textcolor[rgb]{ 1,  0,  0}{55.00} & 65.00 & 90.12 & 65.57 & 78.76 \\
    DC2 & 42.07 & 59.23 & 57.93 & 60.01 & 89.60 & 60.63 & 76.80 & 37.72 & 54.67 & 62.28 & 77.08 & \textcolor[rgb]{ 1,  0,  0}{96.58} & 77.06 & 59.15 \\
    DC3 & 36.20 & 53.15 & 63.20 & 65.02 & 71.19 & 65.00 & 69.36 & 37.82 & 54.86 & 62.18 & 67.43 & 73.16 & 67.40 & 71.35 \\
    DC4 & 38.36 & 55.44 & 61.64 & 62.05 & 80.61 & 63.06 & 73.93 & 40.95 & 58.05 & 59.27 & 79.31 & 80.86 & 79.00 & 80.27 \\
    DC5 & 37.43 & 54.30 & 62.57 & 65.48 & 74.16 & 64.00 & 71.34 & 45.85 & \textcolor[rgb]{ 1,  0,  0}{62.78} & 54.15 & 65.40 & 85.80 & 65.51 & 77.29 \\
    DC6 & 36.41 & 53.37 & 63.59 & 73.70 & 96.05 & 72.80 & 69.47 & 40.73 & 57.67 & 59.27 & 81.15 & 83.18 & 81.56 & 82.37 \\
    DC7 & 37.03 & 53.81 & 62.97 & 61.91 & 63.48 & 62.20 & 63.21 & 32.06 & 47.76 & 67.94 & 55.06 & 56.01 & 56.06 & 55.91 \\
    DC8 & 40.50 & 57.63 & 59.50 & 82.47 & 83.29 & 82.00 & \textcolor[rgb]{ 1,  0,  0}{82.97} & 43.23 & 60.33 & 56.77 & 69.81 & 81.60 & 70.00 & 76.76 \\
    \hline
      & \multicolumn{7}{c}{Block 5 [\%]} \vline &   &   &   &   &   &   &  \\
      \hline
    Data & IoU & Dice & VOE & Sens & Prec & Spec & Acc &   &   &   &   &   &   &  \\
    DC1 & 43.50 & 60.63 & 56.50 & 60.12 & 89.10 & 60.00 & 76.38 &   &   &   &   &   &   &  \\
    DC2 & 42.46 & 59.61 & 57.54 & 61.35 & 91.83 & 60.55 & 77.50 &   &   &   &   &   &   &  \\
    DC3 & 38.15 & 55.23 & 61.85 & 70.00 & 78.42 & 70.15 & 75.43 &   &   &   &   &   &   &  \\
    DC4 & 37.71 & 54.61 & 62.29 & 67.51 & 72.47 & 66.51 & 65.56 &   &   &   &   &   &   &  \\
    DC5 & 38.88 & 55.75 & 61.12 & \textcolor[rgb]{ 1,  0,  0}{84.80} & 91.16 & \textcolor[rgb]{ 1,  0,  0}{84.60} & 72.45 &   &   &   &   &   &   &  \\
    DC6 & 37.07 & 53.96 & 62.93 & 69.56 & 75.92 & 69.65 & 73.75 &   &   &   &   &   &   &  \\
    DC7 & 35.08 & 53.81 & 64.92 & 82.38 & 82.38 & 82.38 & 59.34 &   &   &   &   &   &   &  \\
    DC8 & 41.14 & 58.27 & 58.86 & 67.80 & 81.14 & 68.79 & 76.02 &   &   &   &   &   &   &  \\
    \hline
    \end{tabular}%
  \label{T1:withoutAug}%
  \end{scriptsize}
\end{table}%

\begin{figure}[htbp]
\centering
\includegraphics[width=0.85\textwidth]{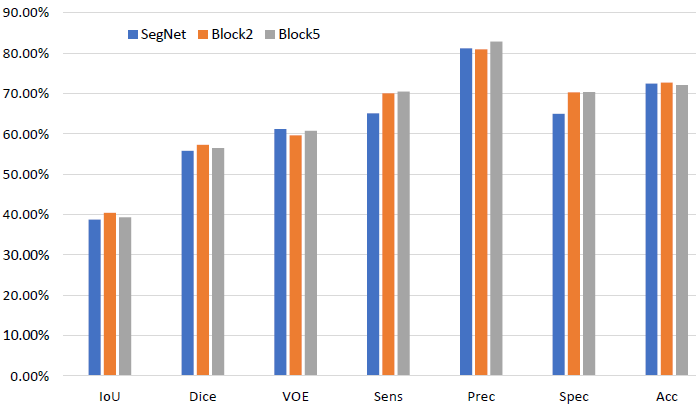}
\caption{Comparison between PDLF-Net (when concatenation is applied on block 2 and block 5) and the original SegNet on small weakly visible dataset (using validation set)}
\label{graph:WithoutAug}
\end{figure}

From table \ref{T1:withoutAug}, the application of pairwise features show improvement of the segmentation results. Block 5 and Block 2 results of the PDLF-Net are presented because they show consistent improvement in all classes compared to other block options. This is because, the deep layers at block 5 (bottom neck layers) in the deep network (PDLF-Net) are responsible for learning specific features of the foreground, therefore adding the joint pairwise features at block 5 emphases more the network to focus on learning the foreground (EM) thus improves the performance. The application of pairwise features on different blocks improves the segmentation performance by 6.21\% acc, 2.9\% IoU, 2.8\% Dice, 14.83\% sens and 15.57\% spec on weak data class 1 (DC1). 6.06\% acc, 1.95\% IoU, 2.08\% Dice, 4.98\% sens, 7.23\% prec and 5.15\% spec on weak data class 3 (DC3). 6.30\% acc, 2.59\% IoU, 2.61\% Dice,  17.26\% sens, 0.25\% prec and 15.94\% spec on DC4. 5.00\% acc, 8.00\% IoU, 8.48\% Dice, 19.32\% sens, 17.00\% prec and 20.6\% spec on DC5. 12.90\% acc, 4.32\% IoU, 4.30\% Dice, 7.45\% sens and 8.76\% spec on DC6. 2.73\% IoU and  2.70\% Dice on DC8. (The comparison above is obtained by taking the original SegNet result for a particular dataset class as a reference and compare it with maximum value of the PDLF-Net result of any block in that particular data class). The average performance results of  the original SegNet and PDLF-Net at block 2 and 5  on all classes are given in figure \ref{graph:WithoutAug} (This is obtained by averaging the results of all classes on a particular method separately and drawing the performance chart for each method).


From the general figure \ref{graph:WithoutAug}, even though the number of dataset for training is very small the PDLF-Net shows improvement in IoU, Dice, VoE, Sens, Prec, Spec and Acc by about 1.66\%, 1.50\%, 1.55\%, 5.34\%, 1.65\%, 5.37\% and 0.28\% respectively. Generally the PDLNet shows improvement, however, the individual errors (VOE) are still high as shown in table \ref{T1:withoutAug}. This is due to the small dataset which cause the networks not to generalize well during training. In order to reduce such errors and increase segmentation performance more, we apply augmentation on all weakly visible dataset and their corresponding GT images.

\subsection{Evaluation of the PDLF-Net on Augmented Dataset}

In order to enhance the performance of PDLF-Net, we augment all the weakly visible EMs and their GT images. Then joint pairwise features are extracted from each image and concatenated to different blocks. Each block is trained and tested independently for each dataset class. Table \ref{T1:withAug} shows the results of the most performing network concatenation configurations.


\begin{table}[h]
 \begin{scriptsize}
  \centering
  \caption{Segmentation results of the original SegNet and PDLF-Net while concatenation is at block 2,3 , 4 and 5, on validation augmented dataset. The results for each weakly visible class 1 to 8 (DC1 - DC8) are given separately in each row. Red coloured values are maximum/best achieved values in each metric}
    \begin{tabular}{|p{0.38cm}|p{0.35cm}p{0.35cm}p{0.35cm}p{0.35cm}p{0.35cm}p{0.35cm}p{0.46cm} |p{0.35cm}p{0.35cm}p{0.35cm}p{0.35cm}p{0.35cm}p{0.35cm}p{0.5cm}|}
    \hline
      & \multicolumn{7}{c}{SegNet [\%]}\vline & \multicolumn{7}{c}{Block 2 [\%]} \vline\\
      \hline
    Data & IoU & Dice & VOE & Sens & Prec & Spec & Acc & IoU & Dice & VOE & Sens & Prec & Spec & Acc \\
    DC1 & 63.97 & 77.92 & 36.03 & 91.25 & 93.91 & 91.26 & 92.16 & 63.95 & 77.94 & 36.05 & 85.15 & 97.06 & 85.20 & 91.26 \\
    DC2 & 56.93 & 72.55 & 43.07 & 73.92 & 96.57 & 73.90 & 85.65 & 59.32 & 74.44 & 40.68 & 77.00 & 96.59 & 77.10 & 87.16 \\
    DC3 & 57.54 & 72.92 & 42.46 & 82.00 & 85.07 & 82.42 & 83.99 & 54.91 & 70.64 & 45.09 & 81.38 & 81.70 & 81.48 & 81.55 \\
    DC4 & 58.74 & 73.78 & 41.26 & 85.00 & 85.29 & 85.14 & 85.29 & 65.45 & 79.04 & 34.55 & 88.32 & 88.34 & 88.00 & 89.74 \\
    DC5 & 64.85 & 78.64 & 35.15 & 91.00 & 90.80 & 90.71 & 91.72 & 63.48 & 77.60 & 36.52 & 87.90 & 92.30 & 86.96 & 89.81 \\
    DC6 & 68.78 & 81.48 & 31.22 & 95.80 & 95.90 & 95.00 & \textcolor[rgb]{ 1,  0,  0}{95.86} & 69.11 & 81.70 & 30.89 & 95.65 & 96.77 & \textcolor[rgb]{ 1,  0,  0}{96.56} & 95.57 \\
    DC7 & 60.43 & 75.21 & 39.57 & 86.01 & 87.38 & 85.30 & 87.39 & 63.89 & 77.88 & 36.11 & 90.00 & 91.57 & 90.57 & 90.58 \\
    DC8 & 57.09 & 72.31 & 42.91 & 83.83 & 82.05 & 82.90 & 83.83 & 62.18 & 76.47 & 37.82 & 88.97 & 89.97 & 87.81 & 88.98 \\
    \hline
      & \multicolumn{7}{c}{Block 3 [\%]}\vline & \multicolumn{7}{c}{Block 4 [\%]} \vline\\
      \hline
      & IoU & Dice & VOE & Sens & Prec & Spec & Acc & IoU & Dice & VOE & Sens & Prec & Spec & Acc \\
    DC1 & 66.28 & 79.71 & 33.72 & 94.21 & 94.30 & 94.22 & 94.26 & 67.57 & 80.59 & 32.43 & 80.43 & 95.87 & 86.46 & 91.60 \\
    DC2 & 61.21 & 75.88 & 38.79 & 73.40 & 96.40 & 73.34 & 85.08 & 59.73 & 74.76 & 40.27 & 76.00 & 96.86 & 75.93 & 86.73 \\
    DC3 & 58.48 & 73.72 & 41.52 & 86.80 & 86.43 & 86.18 & 86.33 & 57.71 & 73.08 & 42.29 & 84.54 & 84.79 & 84.00 & 84.64 \\
    DC4 & 61.53 & 76.02 & 38.47 & 88.00 & 87.00 & 86.98 & 86.98 & 62.02 & 76.35 & 37.98 & 86.10 & 87.33 & 87.22 & 87.98 \\
    DC5 & 62.19 & 76.54 & 37.81 & 90.74 & 90.75 & 90.74 & 88.79 & 68.03 & 80.88 & 31.97 & 91.44 & 91.44 & 91.42 & 90.69 \\
    DC6 & \textcolor[rgb]{ 1,  0,  0}{69.31} & \textcolor[rgb]{ 1,  0,  0}{81.85} & \textcolor[rgb]{ 1,  0,  0}{30.69} & 95.60 & 95.60 & 95.90 & 94.72 & 60.27 & 75.18 & 39.73 & 73.00 & 95.01 & 72.56 & 84.19 \\
    DC7 & 61.14 & 75.74 & 38.86 & 87.95 & 88.01 & 87.90 & 87.96 & 59.50 & 74.26 & 40.50 & 86.46 & 86.47 & 86.40 & 86.47 \\
    DC8 & 62.68 & 76.77 & 37.32 & 88.06 & 88.86 & 88.00 & 88.09 & 61.54 & 75.94 & 38.46 & 87.53 & 88.54 & 87.50 & 87.54 \\
    \hline
      & \multicolumn{7}{c}{Block 5 [\%]} \vline &   &   &   &   &   &   &  \\
      \hline
      & IoU & Dice & VOE & Sens & Prec & Spec & Acc &   &   &   &   &   &   &  \\
    DC1 & 68.08 & 80.82 & 31.92 & 94.10 & 94.16 & 94.00 & 92.43 &   &   &   &   &   &   &  \\
    DC2 & 60.53 & 75.37 & 39.47 & 75.50 & 96.80 & 75.50 & 86.36 &   &   &   &   &   &   &  \\
    DC3 & 57.12 & 72.62 & 42.88 & 78.33 & 86.56 & 79.57 & 82.82 &   &   &   &   &   &   &  \\
    DC4 & 64.19 & 78.04 & 35.81 & 87.00 & 87.12 & 86.45 & 86.77 &   &   &   &   &   &   &  \\
    DC5 & 61.95 & 76.64 & 38.05 & 85.95 & 90.93 & 86.00 & 88.71 &   &   &   &   &   &   &  \\
    DC6 & 68.81 & 81.49 & 31.19 & \textcolor[rgb]{ 1,  0,  0}{96.10} & \textcolor[rgb]{ 1,  0,  0}{97.90} & 95.11 & 95.32 &   &   &   &   &   &   &  \\
    DC7 & 63.19 & 77.37 & 36.81 & 90.90 & 90.89 & 89.89 & 90.42 &   &   &   &   &   &   &  \\
    DC8 & 62.18 & 76.48 & 37.82 & 88.87 & 88.89 & 88.87 & 88.87 &   &   &   &   &   &   &  \\
    \hline
    \end{tabular}%
  \label{T1:withAug}%
  
  \end{scriptsize}
\end{table}%

\begin{figure}[h]
\centering
\includegraphics[width=0.90\textwidth]{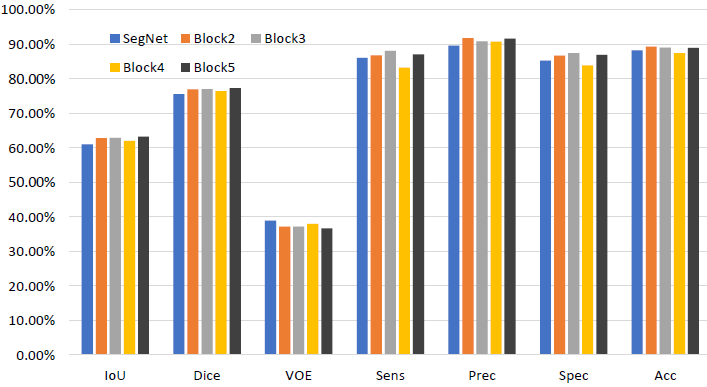}
\caption{The comparison between average performance of the PDLF-Net (when concatenation is applied on block 2,3,4 and block 5) and the original SegNet on all classes of augmented weakly visible EMs. Using validation sets}
\label{graph:WithtAug}
\end{figure}

From table \ref{T1:withAug}, there is an overall improvement of segmentation performance contributed by all blocks. Comparing with the original SegNet, the PDLF-Net shows improvement by an increase of 2.1\% acc, 4.11\% IoU, 2.90\% dice, 2.96\% sens, 3.15\% prec and 2.96\% spec on weakly visible data class 1 (DC1). 1.51\% acc, 4.28\% IoU, 3.33\% dice, 3.08\% sens, 0.29\% prec and 3.20\% spec on DC2. 2.34\% acc, 0.94\% IoU, 0.8\% dice, 4.80\% sens, 1.49\% prec and 3.76\% spec on DC3. 4.45\% acc, 6.71\% IoU, 5.26\% dice, 3.32\% sens, 3.05\% prec and 2.86\% spec on DC4. 3.18\% IoU, 2.24\% dice, 0.44\% sens, 1.5\% prec and 0.71\% spec on DC5. 0.53\% IoU, 0.37\% dice, 0.30\% sens, 2.00\% prec and 1.56\% spec on DC6. 3.19\% acc, 3.46\% IoU, 2.67\% dice, 4.89\% sens, 4.19\% prec and 5.27\% spec on DC7. 5.15\% acc, 5.59\% IoU, 4.46\% dice, 5.18\% sens, 7.92\% prec and 5.97\% spec on DC8. The average performances on all dataset for original SegNet and PDLF-Net are given in figure \ref{graph:WithtAug}.

\begin{figure}
 \centering
  \begin{subfigure}[t]{0.154\textwidth}
  \includegraphics[width=1\textwidth]{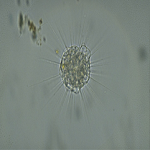}
 \end{subfigure}
 \hspace{0.00cm}
 \begin{subfigure}[t]{0.154\textwidth}
  \includegraphics[width=1\textwidth]{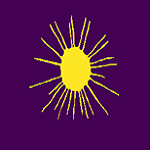}
 \end{subfigure}
 \hspace{0.00cm}
 \begin{subfigure}[t]{0.154\textwidth}
  \includegraphics[width=1\textwidth]{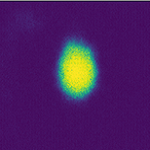}
 \end{subfigure}
  \hspace{0.00cm}
 \begin{subfigure}[t]{0.154\textwidth}
  \includegraphics[width=1\textwidth]{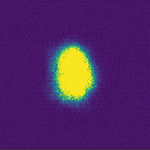}
 \end{subfigure}
  \begin{subfigure}[t]{0.154\textwidth}
  \includegraphics[width=1\textwidth]{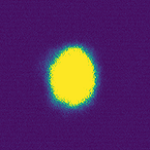}
 \end{subfigure}
 \begin{subfigure}[t]{0.154\textwidth}
  \includegraphics[width=1\textwidth]{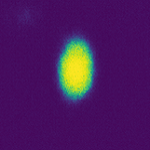}
 \end{subfigure}
  
\vspace{0.5cm}
\begin{subfigure}[t]{0.154\textwidth}
  \includegraphics[width=1\textwidth]{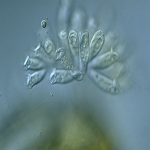}
 \end{subfigure}
 \hspace{0.00cm}
 \begin{subfigure}[t]{0.154\textwidth}
  \includegraphics[width=1\textwidth]{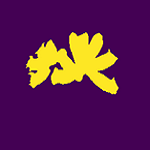}
 \end{subfigure}
 \hspace{0.00cm}
 \begin{subfigure}[t]{0.154\textwidth}
  \includegraphics[width=1\textwidth]{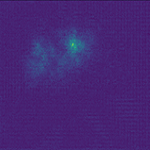}
 \end{subfigure}
  \hspace{0.00cm}
 \begin{subfigure}[t]{0.154\textwidth}
  \includegraphics[width=1\textwidth]{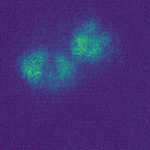}
 \end{subfigure}
  \begin{subfigure}[t]{0.154\textwidth}
  \includegraphics[width=1\textwidth]{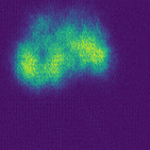}
 \end{subfigure}
 \begin{subfigure}[t]{0.154\textwidth}
  \includegraphics[width=1\textwidth]{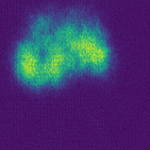}
 \end{subfigure}
 
\vspace{0.5cm}
\begin{subfigure}[t]{0.154\textwidth}
  \includegraphics[width=1\textwidth]{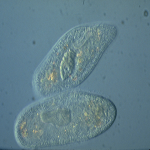}
 \end{subfigure}
 \hspace{0.00cm}
 \begin{subfigure}[t]{0.154\textwidth}
  \includegraphics[width=1\textwidth]{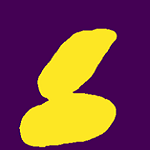}
 \end{subfigure}
 \hspace{0.00cm}
 \begin{subfigure}[t]{0.154\textwidth}
  \includegraphics[width=1\textwidth]{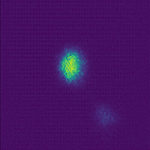}
 \end{subfigure}
  \hspace{0.00cm}
 \begin{subfigure}[t]{0.154\textwidth}
  \includegraphics[width=1\textwidth]{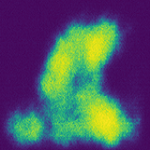}
 \end{subfigure}
  \begin{subfigure}[t]{0.154\textwidth}
  \includegraphics[width=1\textwidth]{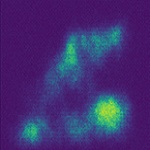}
 \end{subfigure}
 \begin{subfigure}[t]{0.154\textwidth}
  \includegraphics[width=1\textwidth]{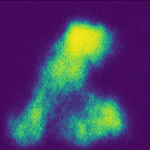}
 \end{subfigure}
 
\vspace{0.5cm}
\begin{subfigure}[t]{0.154\textwidth}
  \includegraphics[width=1\textwidth]{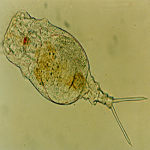}
 \end{subfigure}
 \hspace{0.00cm}
 \begin{subfigure}[t]{0.154\textwidth}
  \includegraphics[width=1\textwidth]{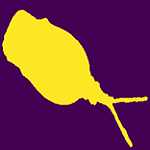}
 \end{subfigure}
 \hspace{0.00cm}
 \begin{subfigure}[t]{0.154\textwidth}
  \includegraphics[width=1\textwidth]{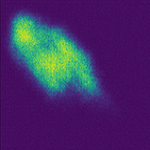}
 \end{subfigure}
  \hspace{0.00cm}
 \begin{subfigure}[t]{0.154\textwidth}
  \includegraphics[width=1\textwidth]{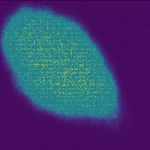}
 \end{subfigure}
  \begin{subfigure}[t]{0.154\textwidth}
  \includegraphics[width=1\textwidth]{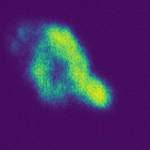}
 \end{subfigure}
 \begin{subfigure}[t]{0.154\textwidth}
  \includegraphics[width=1\textwidth]{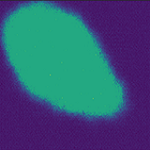}
 \end{subfigure}
 
\vspace{0.5cm} 
\begin{subfigure}[t]{0.154\textwidth}
  \includegraphics[width=1\textwidth]{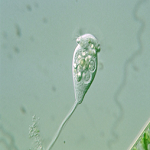}
 \end{subfigure}
 \hspace{0.00cm}
 \begin{subfigure}[t]{0.154\textwidth}
  \includegraphics[width=1\textwidth]{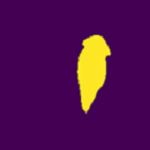}
 \end{subfigure}
 \hspace{0.00cm}
 \begin{subfigure}[t]{0.154\textwidth}
  \includegraphics[width=1\textwidth]{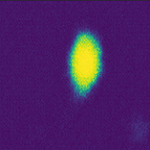}
 \end{subfigure}
  \hspace{0.00cm}
 \begin{subfigure}[t]{0.154\textwidth}
  \includegraphics[width=1\textwidth]{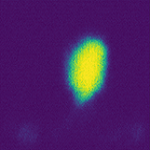}
 \end{subfigure}
  \begin{subfigure}[t]{0.154\textwidth}
  \includegraphics[width=1\textwidth]{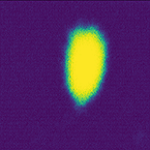}
 \end{subfigure}
 \begin{subfigure}[t]{0.154\textwidth}
  \includegraphics[width=1\textwidth]{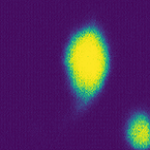}
 \end{subfigure}
 
\vspace{0.5cm} 
 \begin{subfigure}[t]{0.154\textwidth}
  \includegraphics[width=1\textwidth]{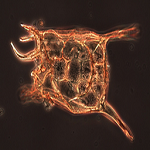}
 \end{subfigure}
 \hspace{0.00cm}
 \begin{subfigure}[t]{0.154\textwidth}
  \includegraphics[width=1\textwidth]{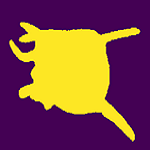}
 \end{subfigure}
 \hspace{0.00cm}
 \begin{subfigure}[t]{0.154\textwidth}
  \includegraphics[width=1\textwidth]{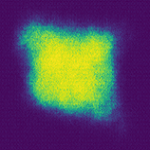}
 \end{subfigure}
  \hspace{0.00cm}
 \begin{subfigure}[t]{0.154\textwidth}
  \includegraphics[width=1\textwidth]{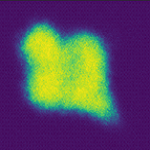}
 \end{subfigure}
  \begin{subfigure}[t]{0.154\textwidth}
  \includegraphics[width=1\textwidth]{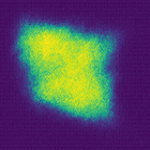}
 \end{subfigure}
 \begin{subfigure}[t]{0.154\textwidth}
  \includegraphics[width=1\textwidth]{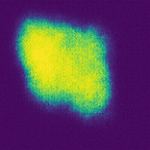}
 \end{subfigure}
 
\vspace{0.5cm}
  \begin{subfigure}[t]{0.154\textwidth}
  \includegraphics[width=1\textwidth]{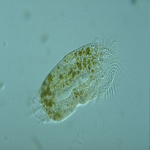}
 \caption{}
 \end{subfigure}
 \hspace{0.00cm}
 \begin{subfigure}[t]{0.154\textwidth}
  \includegraphics[width=1\textwidth]{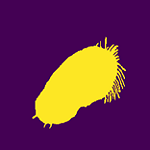}
\caption{}
 \end{subfigure}
 \hspace{0.00cm}
 \begin{subfigure}[t]{0.154\textwidth}
  \includegraphics[width=1\textwidth]{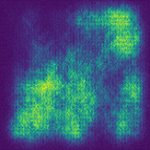}
 \caption{}
 \end{subfigure}
  \hspace{0.00cm}
 \begin{subfigure}[t]{0.154\textwidth}
  \includegraphics[width=1\textwidth]{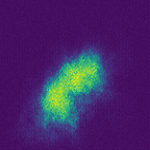}
 \caption{}
 \end{subfigure}
  \begin{subfigure}[t]{0.154\textwidth}
  \includegraphics[width=1\textwidth]{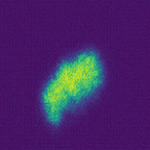}
 \caption{}
 \end{subfigure}
 \begin{subfigure}[t]{0.154\textwidth}
  \includegraphics[width=1\textwidth]{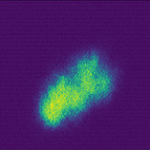}
 \caption{}
 \end{subfigure}

\caption{Example of segmented images on weakly visible dataset. column (a) are original images, (b) ground  truth images, (c) segmented images by original SegNet, (d) segmented images by PDLF-Net when concatenation is at block 2 (e) concatenation at block 3 or 4, (f) concatenation at block 5. Form top to bottom are different classes, the top most row is of weakly data class 1 (DC1), second row DC3, followed by DC4, DC5, DC6, DC7 and DC8 respectively.} 
\label{segmented images}
\end{figure}

Figure \ref{graph:WithtAug} shows that the average improvement of about 1.09\% acc, 2.20\% IoU, 1.75\% dice, 2.00\% sens, 2.17\% prec and 2.15\% spec is observed on segmentation using PDLF-Net compared to the original SegNet. The overall average maximum results achieved by the PDLF-Net are 89.33\%, 63.26\%, 77.35\%, 36.74\%, 88.10\%, 91.79\% and 87.48\% by acc, IoU, Dice, VOE, sens, prec and spec  respectively. Moreover, the visual comparison of segmented images on original SegNet and PDLF-Net at blocks 2, 3, 4 and 5  are given in figure \ref{segmented images}.


Comparing the observation performance from figure \ref{segmented images}, the PDLF-Net shows better segmentation results. For instance, in data class DC3 and DC4 ($2^{nd}$ and $3^{rd}$ rows from the top) SegNet in (c) has not been able to show the foreground while there is a good segmented output of  the same image by PDLF-Net in (e) and (f). In DC8 (last row), SegNet over-segments the image while good visual results are observed by PDLF-Net when concatenation of pairwise feature is at block 2, 3 and 5. Generally, the visual results show great improvement of segmentation results when using PDLF-Net.

\subsection{Evaluation of the PDLF-Net on Test Dataset} 
\begin{figure}[H]
\centering
\includegraphics[width=0.85\textwidth]{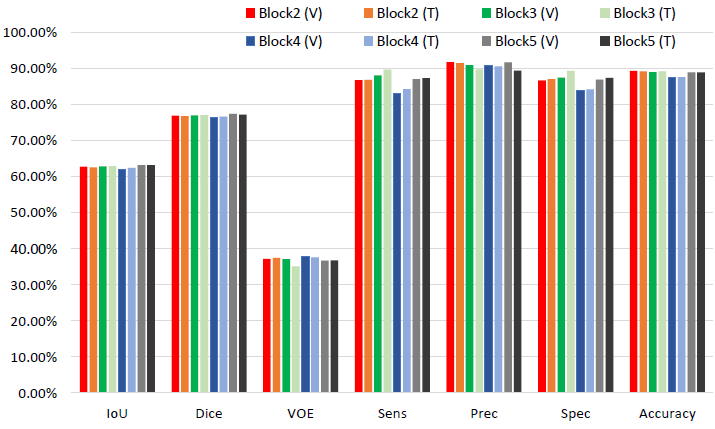}
\caption{The segmentation performance of PDLF-Net on both test set and validation set. Block 2(V), Block 3(V), Block 4(V) and Block 5(V) are results for validation sets respectively. Block 2(T), Block 3(T), Block 4(T) and Block 5(T) are results for test sets respectively }
\label{graph:TestAug}
\end{figure}

To evaluate more the effectiveness of the PDLF-Net, we examine it on the test dataset. The test dataset contains 480 images, which are twice in number to the training and validation sets. The average segmentation performance of the PDLF-Net  on test dataset for all classes is shown in figure \ref{graph:TestAug}. The graph shows a comparison of both the test set and validation set for each block performance.

From figure \ref{graph:TestAug}, each pair of bars from left to right, are of similar configuration (blocks) applied on validation and test sets respectively. The performance of the PDLF-Net is almost similar in both validation and test sets although the number of test dataset is twice. This shows the great effectiveness of the PDLF-Net on unseen dataset (test set). The highest average performances of PDLF-Net on test dataset are 89.24\% accuracy, 63.20\% IoU, 77.27\% Dice, 35.15\% VOE, 89.72\% sensitivity, 91.44\% precision and 89.30\% specificity .

\subsubsection{Evaluation of the model performance on more challenging Test Datasets}

\begin{figure}[htbp]
\centering
\includegraphics[width=0.85\textwidth]{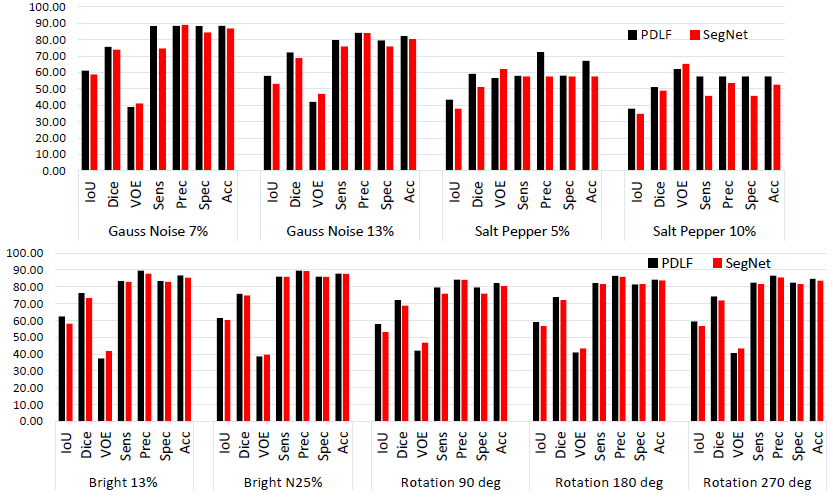}
\caption{Performance of the PDLF-Net on test EMs images which have been subjected to rotation of 90,180 and 270 degrees. Additional of Gaussian noise by 7\% and 13\%. Additional of  salt-pepper noise by 5\% and 10\%. Illumination change by increasing brightness by 13 \% (Bright 13\%) and decreasing by 25\% (Bright N25\%).}
\label{graph:ChallengingDataGRAPH}
\end{figure}

In order to evaluate the performance of the PDLF-Net on more challenging dataset, we test it on images which have been subjected to rotation, illumination change and additional noise as indicated in the figure \ref{graph:ChallengingDataGRAPH}. To observe the improved capability of the PDLF-Net on learning image features on challenging dataset, we compare it with the base model SegNet. 


It can be observed from the graphs on figure \ref{graph:ChallengingDataGRAPH} that the PDLF-Net performs better on all challenging images with an average improved performance of more than 2.00\% against SegNet on each metric. This justify the capability of the PDLF-Net to capture more spatial features in noisy, transparent and low contrast images.

\subsubsection{Training and Testing Time Evaluation}

In this section we compare the training and testing time of the PDLF-Net against other well-known CNN based segmentation models as shown in  Table \ref{tab:ModelComplexity}.

\begin{table}[htbp]
\begin{scriptsize}

  \centering
  \caption{Average training and testing time for the PDLF-Net (when concatenation is at Block2, 3, 4 and 5), SegNet, Unet and Fully connected network (FCN)}
  
    \begin{tabular}{|llllllll|}
    \hline
    Model & Block2 & Block3 & Block4 & Block 5 & SegNet & Unet & FCN \\
    \hline
    Training time (min) & 12.17 & 11.98 & 11.18 & 11.08 & 10.53 & 7.78 & 9.56 \\
    Testing time (sec) & 6.19 & 5.92 & 6.15 & 6.12 & 5.92 & 3.21 & 4.31 \\
    \hline
    \end{tabular}%
  \label{tab:ModelComplexity}%
  \end{scriptsize}
\end{table}%

From table \ref {tab:ModelComplexity}, although the training and testing times for the PDLF-Net are a bit higher compared to other models, they are generally still low and feasible for practical segmentation tasks.

\subsubsection{Comparison of the PDLF-Net Against Other State-of-the Art Segmentation Networks}
We conduct comparison tests of the proposed model against U-net, FCN, SegNet, Canny edge based segmentation, Otsu thresholding, $k$-means clustering and region growing segmentation techniques on the same test dataset. Because the classical methods (Canny, Otsu and region growing) need post-processing to have better segmentation results, we use same post-processing techniques for all so as to unify the results. During test experiments PDLF-Net, SegNet, Unet and FCN are all trained using augmented training EMs dataset and tested on the same test dataset. The classical methods are subjected to test datasets only. The results obtained for each  networks are represented in table \ref{tab:TestResulALL}.

\begin{table}[htbp]
\begin{scriptsize}

  \centering
  \caption{The test results of the PDLF-Net (when concatenation is at block 2 (BL2), BL3, BL4 and BL5), SegNet, U-net, FCN, Otsu, Canny, $k$-means and region growing (RG).}
        \begin{tabular}{|p{0.54cm}p{0.54cm}p{0.54cm}p{0.53cm}p{0.53cm}p{0.53cm}p{0.53cm}p{0.54cm}p{1.05cm}p{0.54cm}p{0.54cm}p{0.54cm}|}
        \hline
      & SegNet & BL2 & BL3 & BL4 & BL5 & Unet & FCN & $k$-means & Canny & Otsu & RG \\
      \hline
    IoU & 61.02 & 62.56 & 62.91 & 62.41 & \textcolor[rgb]{ 1,  0,  0}{63.20} & 60.65 & 36.85 & 29.52 & 37.65 & 38.25 & 31.08 \\
    Dice & 75.58 & 76.79 & 77.08 & 76.65 & \textcolor[rgb]{ 1,  0,  0}{77.27} & 73.06 & 53.84 & 36.15 & 49.45 & 47.65 & 41.57 \\
    VOE & 38.98 & 37.44 & \textcolor[rgb]{ 1,  0,  0}{35.15} & 37.59 & 36.80 & 39.35 & 63.15 & 63.84 & 50.54 & 52.34 & 61.42 \\
    Sens & 86.00 & 86.88 & \textcolor[rgb]{ 1,  0,  0}{89.72} & 84.30 & 87.35 & 76.12 & 65.00 & 60.86 & 76.21 & 70.37 & 51.52 \\
    Prec & 89.88 & \textcolor[rgb]{ 1,  0,  0}{91.44} & 89.79 & 90.57 & 89.40 & 81.59 & 65.33 & 34.22 & 39.90 & 50.34 & 51.95 \\
    Spec & 86.04 & 87.06 & \textcolor[rgb]{ 1,  0,  0}{89.30} & 84.20 & 87.37 & 87.00 & 65.34 & 66.58 & 64.66 & 69.31 & 68.23 \\
    Acc & 88.10 & \textcolor[rgb]{ 1,  0,  0}{89.24} & 89.21 & 87.60 & 88.92 & 84.94 & 65.37 & 65.33 & 68.24 & 68.45 & 64.64 \\
    \hline
    \end{tabular}%
  \label{tab:TestResulALL}%
  \end{scriptsize}
  
\end{table}%

It can be observed from the table \ref{tab:TestResulALL} that the PDLF-Net performs better than other networks by having the highest values in all metrics. The average good performing blocks for the PDLF-Net are block 3 and block 5.

\subsection{Method's Limitations}
It should be noted that, although the proposed method has shown potential on EMs, it focuses only on segmentation of one (single) or two   microorganisms on the image and not biofilms. Example of the segmentation of two EMs on the same image can be seen in Fig. \ref{segmented images} in classes DC3 and DC4, while other classes contain only one EM on every image. In the future work, we will extend our scope to testing our novel method on microorganim dataset with more than two microorganisms, clusters, and biofilms.

\section{Conclusion and Future Work}
\label{s:Conc}
In this research we propose a \textit{Pairwise Deep Learning Feature Network} for segmentation of weakly visible EMs. It combines the advantages of both hand crafted features (by identifying the Shi and Tomas interest points of the foreground ) and deep learning features (by extracting deep learning features on the patches which are centered on each interest point). Then, in order to learn the intermediate spatial characteristics between the nearby interest points, we  pair the extracted deep learning features using the Delaunay triangulation theorem. The results show that the proposed network upon improving the performance of the base mode SegNet, it can focus more on the foreground which can overcome the segmentation challenges on image such as noise and low contrast. Apart from being useful in segmentation of EMs, the proposed network can find more application in segmentation of brain tumor and breast cancer images. 

During initial  experiments we tested the pairwise deep learning features on binary classification of two EMs classes using SVM. Promising results were obtained. Therefore, the pairwise features can also be suitable not only in segmentation tasks but also in classification and image matching works.

In the future work, we plan to use other superior convolution neural networks such as Inception, Xception and DenseNet for extraction of deep learning features to improve more our segmentation results.

\section{Acknowledgements}
\label{S:ack} 

We thank Prof. B. Zhou, Dr. F. Ma (University of Science and Technology Beijing, China), 
Prof. Y. Zou (Freiburg University, Germany), 
B.E. X. Zhu (Johns Hopkins University, US) 
and B.E. B. Lu (Huazhong University of Science and Technology, China) for their previous cooperations in this work. We also thank 
Miss Z. Li and Mr. G. Li, for their important discussion. 
This work is supported by ``National Natural Science Foundation of China'' (No. 61806047).

\begin{scriptsize}

\bibliography{mybibfile}
\end{scriptsize}

\end{document}